\documentclass[11pt]{article}

\usepackage[final]{acl}

\usepackage{times}
\usepackage{latexsym}

\usepackage[T1]{fontenc}

\usepackage[utf8]{inputenc}

\usepackage{microtype}

\usepackage{inconsolata}

\usepackage{graphicx}

\usepackage{amsmath}
\usepackage{hyperref}
\usepackage{bbm}

\title{mucAI at WojoodNER 2024: Arabic Named Entity Recognition with Nearest Neighbor Search}

\author{Ahmed Abdou \\
  Technical University of Munich, Germany \\
  \texttt{ahmed.abdou@tum.de} \\\And
  Tasneem Mohsen \\
  Helwan University of Cairo, Egypt \\
  \texttt{tasneem\_20210239@fci.helwan.edu.eg} \\}

\begin{document}
\maketitle
\begin{abstract}
Named Entity Recognition (NER) is a task in Natural Language Processing (NLP) that aims to identify and classify entities in text into predefined categories.
However, when applied to Arabic data, NER encounters unique challenges stemming from the language's rich morphological inflections, absence of capitalization cues, and spelling variants, where a single word can comprise multiple morphemes.
In this paper, we introduce Arabic KNN-NER, our submission to the Wojood NER Shared Task 2024 (ArabicNLP 2024). We have participated in the shared sub-task 1 Flat NER. In this shared sub-task, we tackle fine-grained flat-entity recognition for Arabic text, where we identify a single main entity and possibly zero or multiple sub-entities for each word.
Arabic KNN-NER augments the probability distribution of a fine-tuned model with another label probability distribution derived from performing a KNN search over the cached training data. Our submission achieved 91\% on the test set on the WojoodFine dataset, placing Arabic KNN-NER on top of the leaderboard for the shared task.

\end{abstract}

\section{Introduction}
Named Entity Recognition (NER) is a downstream task within Natural Language Processing (NLP) that entails identifying the entity type for each word in a given sentence. These entities usually belong to predefined tags such as persons, locations, or dates. 
NER has proven beneficial for several downstream tasks, including relation extraction, machine translation, co-reference resolution, and information extraction. 
The significance of the NER task has led to the development of various approaches, including span-based classification \cite{span-based}, sequence labeling, and sequence-to-sequence generation. Recently, with the widespread adoption of large-scale pre-trained language models (PLMS), in-context learning-based approaches \cite{chen2023learning} have also emerged as a prominent method.

However, applying NER to Arabic data presents additional challenges. Unlike English, Arabic does not use capital letters at the beginning of words, making it more challenging to identify nouns and determine the start and end of entities. Multiple word variants can also refer to the same semantic meaning \cite{qu2023survey}. Furthermore, Arabic is rich in morphological inflections, meaning that a single word can consist of multiple morphemes \cite{benajiba2008arabic}. Finally, annotated Arabic corpora for the NER task are limited compared to those available for English \cite{qu2023survey}.

Previous attempts to enrich the Arabic NER corpus as the multilingual dataset ACE \cite{walker2005ace}, ANERCorp \cite{benajiba2007anersys}, and Ontonotes5 \cite{weischedel2013ontonotes}. More recently, the Wojood \cite{jarrar2023wojoodner} dataset was introduced, a large-scale Arabic NER dataset collected from multiple sources covering Modern Standard Arabic (MSA) and dialects. However, all of these datasets are all annotated with coarse-grained entity types \cite{jarrar2023wojoodner}. The most recent corpus is WojoodFine \cite{jarrar-etal-2024-wojoodner}, which extends Wojood by providing 31 fine-grained annotations, introducing subtypes for some of the main entity types.

In this paper, we tackle the shared subtask Flat NER with subtypes using the WojoodFine dataset; we comprehensively describe the data in section \ref{sec-data}. We propose enhancing a fine-tuned NER model by integrating KNN search over the training entities during the inference phase. KNN-NER \cite{knnner} is a framework that can be applied to models that have already been fine-tuned, requiring no additional training or fine-tuning. Applying the KNN-NER framework has achieved a 91\% micro-F1 score and placed us on top of the leaderboard for the shared subtask 1 Flat NER.

\section{Task Definition}
The shared subtask 1 is Flat NER with subtypes. In this task, for a given sentence, the objective is to identify and classify named entities, which may span multiple words. For each identified entity, the task is to determine the main entity type and up to two levels of subtypes, if they exist. For example, in Figure \ref{fig:data_example} (b), the main entity DATE spans two words and has no subtypes. Meanwhile, the entity ORG spans four words and has two levels of subtypes. The first-level subtype is ORG\_FAC (a subtype of ORG), and the second-level subtypes are COM (a subtype of ORG) and BUILDING-OR-GROUND (a subtype of FAC). More formally, given a sequence of tokens $T$ of length $m$ denoted as $T = (t_{1}, t_{2}, ...., t_{m})$, the goal is to identify and output a set of named entities. Each named entity is represented as a tuple $(s, e, main\_tag, [sub\_tags])$, where $s$ and $e$ are the start and end tokens of the entity, respectively. The tuple also includes the main entity tag and up to two levels of subtypes.

\section{Data} 
\label{sec-data}
\begin{figure}[t]
  \includegraphics[width=\columnwidth]{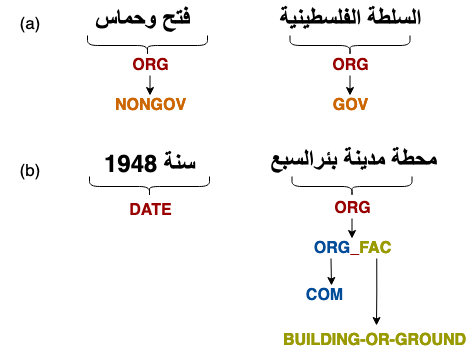}
  \caption{Examples from WojoodFine illustrating (a) same main entities with different subtypes. (b) main entity with zero and multiple subtypes.}
  \label{fig:data_example}
\end{figure}

\begin{figure}[t]
  \includegraphics[width=\columnwidth]{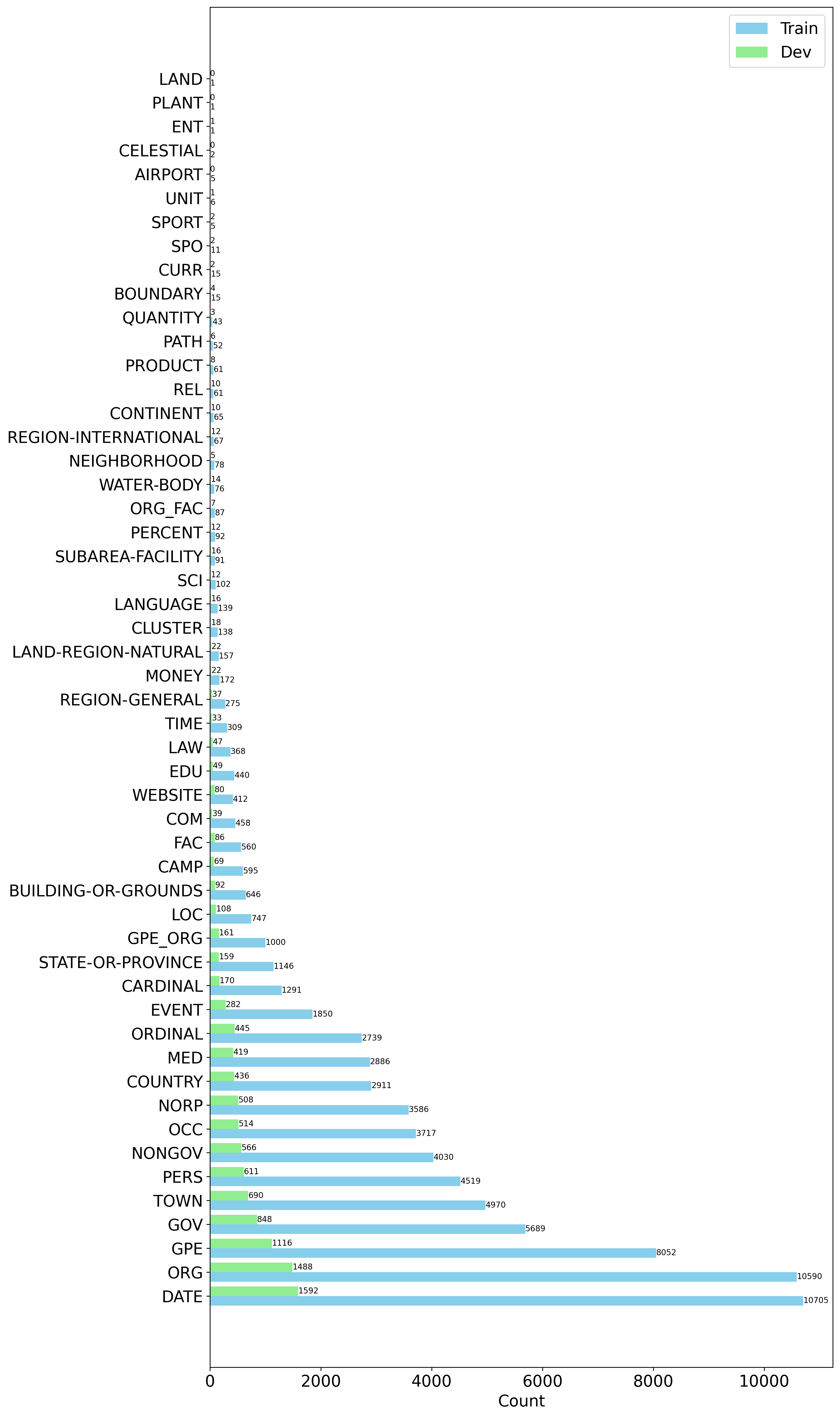}
  \caption{Distribution of NER tags in WojoodFine Subtask1 (i.e., FlatNER) across the training, development splits.}
  \label{fig:entity_dist}
\end{figure}

We conducted our work in WojoodFine dataset \cite{liqreina2023arabic} provided in the shared task \cite{jarrar-etal-2024-wojoodner}. WojoodFine is an extended version of Wojood \cite{jarrar2022wojood} dataset, which is a NER dataset with 550,000 tokens manually annotated across 21 entity types; approximately 80\% of Wojood's data was sourced from Modern Standard Arabic (MSA) articles, while 12\% was gathered from social media content in Palestinian and Lebanese dialects. WojoodFine extends the Wojood dataset by providing fine-grain annotations for entity sub-types. Namely, each of the words that has one of the main entity types (Geopolitical Entity (GPE), Organization (ORG), Location (LOC), and Facility (FAC)) can have from zero to multiple subtypes from the predefined 31 subtypes. For example, in Figure \ref{fig:data_example} (a), words with the same main entity can have different subtypes. The train and development entities distribution can be found in Figure \ref{fig:entity_dist}. For the exact mapping of main entities to subtypes, we refer the reader to check \cite{liqreina2023arabic}.

\section{System Description}
Our approach is centered on fine-tuning a language model based on BERT’s transformer architecture \cite{devlin2018bert}. Our methodology is inspired by \cite{knnner}, which employs a two-step process: joint vanilla fine-tuning followed by KNN at inference time.

\subsection{Joint Finetuning}

\begin{figure}[t]
  \includegraphics[width=\columnwidth]{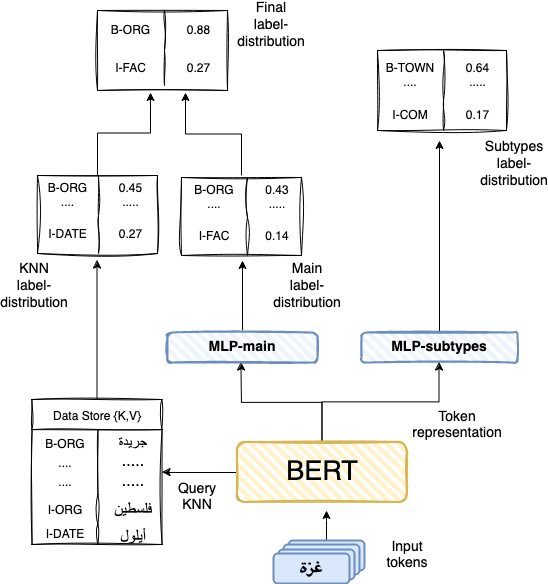}
  \caption{The proposed Model workflow for flat NER with subtypes, jointly fine-tuned with two MLP heads for main entity and subtype prediction. KNN search is applied during inference to enhance prediction accuracy.}
  \label{fig:model}
\end{figure}

The architecture of our solution, illustrated in Figure \ref{fig:model}, uses BERT as the backbone for generating word embeddings. These embeddings are then fed into two MLP heads that are trained jointly. The first head predicts one of the predefined 21 main entity tags. It is designed with 43 output neurons (the outside tag, main tags prefixed with 'I', and main tags prefixed with 'B'). This head is followed by a softmax layer and trained using cross-entropy loss: 
\begin{equation}
P_{main} = Softmax(MLP_{1}(e_{i}))
\end{equation}.
The second head predicts one of the predefined 31 sub-entities. It has 62 output neurons (sub-entities prefixed with 'I' and 'B'). This head is followed by a sigmoid function applied to each output neuron and trained with binary cross-entropy loss, with each neuron's output thresholded at 0.5
\begin{equation}
P_{sub} = Sigmoid(MLP_{2}(e_{i}))
\end{equation}.


\subsection{KNN-NER}
\subsubsection{Datastore Construction}
Post fine-tuning, we obtain contextualized representations $e_i$ for every token in each sentence of the training set using the trained model. The datastore is built by performing a single forward pass over the entire training set. The datastore ${K, V}$ comprises all contextualized representation-entity main-label pairs for each word in each sentence in the training set $D_{t}$, defined as:


\begin{equation}
\resizebox{.8\hsize}{!} 
{$
\{K,V\} = \{(e_i,l_i)| \forall e_i \in s, \forall l_i \in l, (s,l) \in D_{t}\}
$}
\end{equation}

Where $e_i$ is the ith token in the sentence $s$, and $l_i$ is the word corresponding label.

\subsubsection{KNN Inference}
During inference time, we query the datastore using the contextualized representation of every token in each test sentence to find the k-nearest neighbors $N$ according to a similarity score $sim(.,.)$. Then, we derive the distribution of labels $P_{kNN}$ using labels of the retrieved neighbors while aggregating probability mass for each label across all its occurrences in the retrieved neighbors (labels that do not appear in the retrieved $N$ are assigned zero probability). Intuitively, the closer a neighbor is to the test instance, the larger its weight is. Moreover, the higher the number of neighbors having the same label, the higher the probability mass of this label in the derived $P_{kNN}$ probability distribution. More formally,

\begin{equation}
    P_{kNN} \propto \sum_{(k,v)\in N} \mathbbm{1}_{l_i = v}\exp(\frac{sim(e_i, k)}{\tau})
\end{equation}

\begin{equation}
    sim(a, b) = \frac{a \cdot b}{|a||b|}
\end{equation}

Where $\tau$ denotes the temperature hyper-parameter and $sim(a,b)$ is the cosine similarity between two vectors $a$ and $b$. Finally, we interpolate the $P_{main}$ with $P_{kNN}$ with an interpolation factor $\lambda$ as :

\begin{align}
P_{final} =  \lambda P_{main} + (1 -\lambda) P_{kNN}
\label{interpolation}
\end{align}
\section{Results}
\subsection{Experimental Setup}
After reviewing the performance of various solutions and foundational models used in the 2023 WojoodNER task \cite{jarrar2023wojoodner}, we selected AraBERTv02 to be our base model.

All experiments were conducted using a single V100 GPU on Google Colab. We utilized the validation dataset to select the hyperparameters of the model and the KNN search. The maximum input sequence length was set to 512; sequences exceeding this length were truncated, while shorter sequences were padded. Each experiment was run for 20 epochs. We used the AdamW optimizer \cite{loshchilov2017decoupled} with a learning rate of $\eta = 2e{-5}$, an exponential learning rate scheduler with a gamma of $\gamma = 0.95$, a batch size of $B = 16$, and a dropout rate of 0.1. For the KNN search, we perform a grid search by varying the number of retrieved neighbors $N$ and the interpolation factor $\lambda$. Specifically, we explore multiple powers of two for $N$ ranging from $2^3$ to $2^9$ and adjust the interpolation factor from 0 to 1 in 0.1 step. In all inference variants, we set the temperature $\tau$ to 1. All models are implemented using PyTorch, and Huggingface Transformers. The code used for the experiments is available on GitHub\footnote{\url{https://github.com/AhmedAbdel-Aal/WNER_24_sharedtask}}.

\subsection{Results}
We present the micro F1, precision, and recall scores for the development and test sets in Tables \ref{results}, both for vanilla fine-tuning and for using KNN search at inference time. Furthermore, Table \ref{results_teams} highlights our performance in comparison to other teams.

\begin{table}
  \centering
  \begin{tabular}{llll}
    \hline
    \textbf{Model}  & \textbf{P} & \textbf{R}  & \textbf{F1-score} \\
    \hline
    \multicolumn{4}{c}{\textbf{Dev Set}} \\
    \hline
    joint finetuning & 92.47 & 91.24 & 91.87\\
    +KNN & 92.62 & 91.66 & 92.15\\
    \hline
    \multicolumn{4}{c}{\textbf{Test Set}} \\
    \hline
    joint finetuning & 90.23 & 89.95 & 90.00\\
    +KNN & 91.00 & 90.00 & 91.00\\    
    \hline
  \end{tabular}
  \caption{Results on Flat NER}
  \label{results}
\end{table}

\begin{table}
  \centering
  \begin{tabular}{llll}
    \hline
    \textbf{Team}  & \textbf{F1-score} & \textbf{Rank} \\
    \hline
    mucAI (ours) & 91.00 & 1\\    
    \hline
    muNERa & 90.00 & 2 \\
    Addax & 90.00 & 2\\
    Baseline & 89.00 & \\
    DRU - Arab Center & 87.00 & 4 \\
    Bangor & 86.00 & 5\\
    \hline
  \end{tabular}
  \caption{Shared task leaderboard and micro-F1}
  \label{results_teams}
\end{table}
\section{Discussion}
The results from our experiments demonstrate the effectiveness of incorporating KNN search at inference time for the flat NER task in Arabic. The comparison between vanilla fine-tuning without and with KNN search reveals a consistent improvement in the F1-score, as shown in Table \ref{results}; this is analogous to the results presented in \cite{knnner}.

\begin{figure}[t]
  \includegraphics[width=\columnwidth]{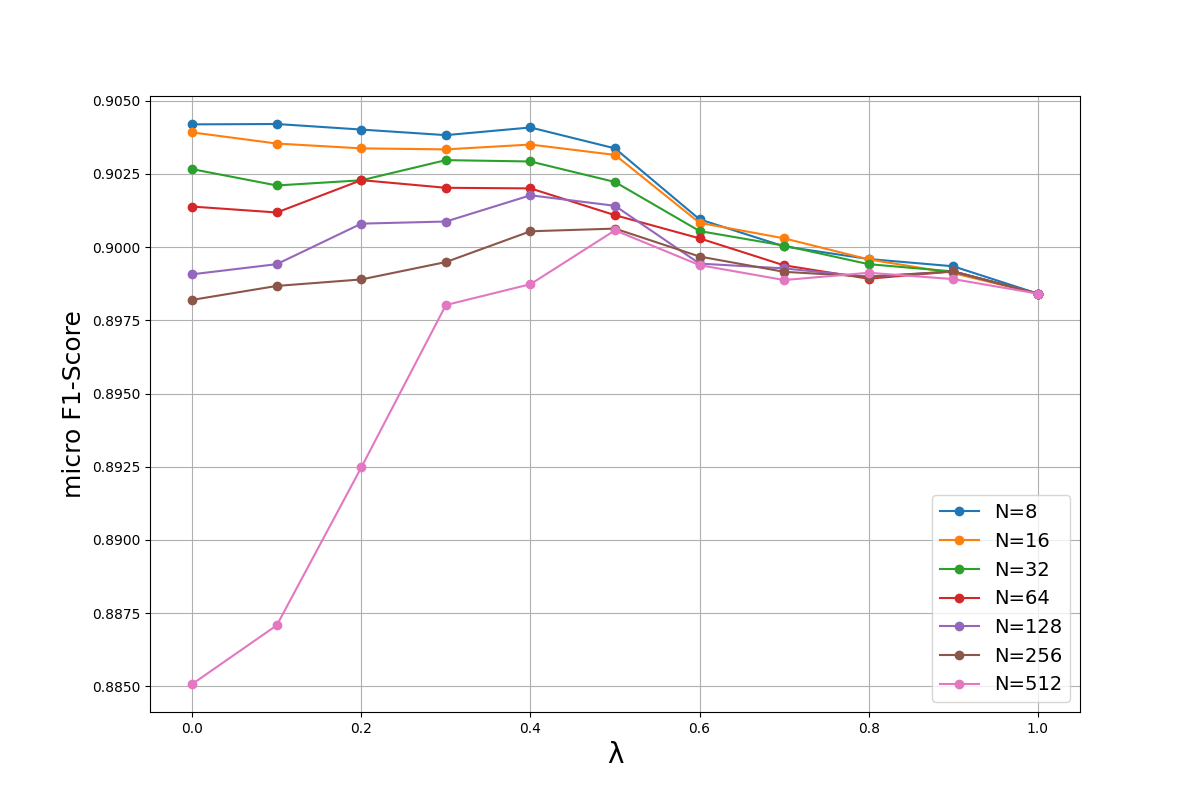}
  \caption{Sensitivity of KNN search to number of neighbors ($N$), and interpolation factor $\lambda$.}
  \label{fig:experiments}
\end{figure}

We show in Figure \ref{fig:experiments} the sensitivity of our system's micro F1-score for different values of $N$ and interpolation factor $\lambda$. If the interpolation factor is One ($\lambda = 1$), the final distribution converges to only the baseline,\begin{equation}
P_{final} =P_{main}
\end{equation}

while if ($\lambda = 0$) the final distribution converges to only use the KNN-RR.
\begin{equation}
P_{final} = P_{kNN}
\end{equation}

Using only KNN-RR appears to be competitive or even better than using the fine-tuned model for small values of N. However, its performance drops to 88.5\% when $N=512$. This decline can be attributed to class imbalance between entity tags and the label O. As the number of retrieved neighbors increases, more neighbors with the label O are retrieved, thereby increasing the probability mass of the label O in $P_{kNN}$.

\section{Limitations and Future Work}

A limitation of this study is that KNN-NER was not evaluated on multiple models, which limits the assessment of its robustness and model-agnostic property of the KNN-NER framework. Another limitation is the increased inference time. Since KNN requires searching for labels in the datastore during inference, the overall inference time is extended by both the model's processing time and the additional time required for the KNN search. This increased inference time may affect the practicality and efficiency of our model in real-world applications. Another limitation is that we did not explore other similarity scores than cosine similarity, nor assessed the quality of the similarity scores it produced. Finally, Future work should include exploring applying KNN-NER to subtypes.

\section{Conclusion}
In this shared task, we tackled flat NER with subtypes on the WojoodFine corpus, where we trained the model jointly with two MLP heads, one for predicting the main entity and the other for predicting possibly multiple subtypes. We finetuned Arabert and applied KNN search over the training set during inference time to enhance the model's capability of predicting the main entity type for each token in the test set. 
The motivation behind incorporating KNN search was to improve the model's performance without requiring any further training after the initial fine-tuning phase. This approach aimed to efficiently utilize the trained model by leveraging KNN over the training set. 
The results show an improvement with the incorporation of KNN search. Specifically, the micro-F1 scores for the development set increased from 91.62 to 92.90, and for the test set from 90.09 to 91.00, indicating a robust enhancement in performance. Our approach ranked first in the shared task leaderboard.

\bibliography{custom}

\begin{thebibliography}{14}
\providecommand{\natexlab}[1]{#1}

\bibitem[{Benajiba and Rosso(2008)}]{benajiba2008arabic}
Yassine Benajiba and Paolo Rosso. 2008.
\newblock Arabic named entity recognition using conditional random fields.
\newblock In \emph{Proc. of Workshop on HLT \& NLP within the Arabic World, LREC}, volume~8, pages 143--153.

\bibitem[{Benajiba et~al.(2007)Benajiba, Rosso, and Bened{\'\i}ruiz}]{benajiba2007anersys}
Yassine Benajiba, Paolo Rosso, and Jos{\'e}~Miguel Bened{\'\i}ruiz. 2007.
\newblock Anersys: An arabic named entity recognition system based on maximum entropy.
\newblock In \emph{Computational Linguistics and Intelligent Text Processing: 8th International Conference, CICLing 2007, Mexico City, Mexico, February 18-24, 2007. Proceedings 8}, pages 143--153. Springer.

\bibitem[{Chen et~al.(2023)Chen, Lu, Lin, Lou, Jia, Dai, Wu, Cao, Han, and Sun}]{chen2023learning}
Jiawei Chen, Yaojie Lu, Hongyu Lin, Jie Lou, Wei Jia, Dai Dai, Hua Wu, Boxi Cao, Xianpei Han, and Le~Sun. 2023.
\newblock Learning in-context learning for named entity recognition.
\newblock \emph{arXiv preprint arXiv:2305.11038}.

\bibitem[{Devlin et~al.(2018)Devlin, Chang, Lee, and Toutanova}]{devlin2018bert}
Jacob Devlin, Ming-Wei Chang, Kenton Lee, and Kristina Toutanova. 2018.
\newblock Bert: Pre-training of deep bidirectional transformers for language understanding.
\newblock \emph{arXiv preprint arXiv:1810.04805}.

\bibitem[{Jarrar et~al.(2023)Jarrar, Abdul-Mageed, Khalilia, Talafha, Elmadany, Hamad, and Omar}]{jarrar2023wojoodner}
Mustafa Jarrar, Muhammad Abdul-Mageed, Mohammed Khalilia, Bashar Talafha, AbdelRahim Elmadany, Nagham Hamad, and Alaa' Omar. 2023.
\newblock Wojoodner 2023: The first arabic named entity recognition shared task.
\newblock \emph{arXiv preprint arXiv:2310.16153}.

\bibitem[{Jarrar et~al.(2024)Jarrar, Hamad, Khalilia, Talafha, and Elmadany}]{jarrar-etal-2024-wojoodner}
Mustafa Jarrar, Nagham Hamad, Mohammed Khalilia, Bashar Talafha, and Muhammad Elmadany, AbdelRahim Abdul-Mageed. 2024.
\newblock {W}ojood{NER} 2024: The second {A}rabic named entity recognition shared task.
\newblock In \emph{Proceedings of the 2nd Arabic Natural Language Processing Conference (Arabic-NLP), Part of the ACL 2024.} Association for Computational Linguistics.

\bibitem[{Jarrar et~al.(2022)Jarrar, Khalilia, and Ghanem}]{jarrar2022wojood}
Mustafa Jarrar, Mohammed Khalilia, and Sana Ghanem. 2022.
\newblock Wojood: Nested arabic named entity corpus and recognition using bert.
\newblock \emph{arXiv preprint arXiv:2205.09651}.

\bibitem[{Liqreina et~al.(2023)Liqreina, Jarrar, Khalilia, El-Shangiti, and Mageed}]{liqreina2023arabic}
Haneen Liqreina, Mustafa Jarrar, Mohammed Khalilia, Ahmed El-Shangiti, and Muhammad~Abdul Mageed. 2023.
\newblock Arabic fine-grained entity recognition.
\newblock In \emph{Proceedings of ArabicNLP 2023}, pages 310--323.

\bibitem[{Loshchilov and Hutter(2017)}]{loshchilov2017decoupled}
Ilya Loshchilov and Frank Hutter. 2017.
\newblock Decoupled weight decay regularization.
\newblock \emph{arXiv preprint arXiv:1711.05101}.

\bibitem[{Qu et~al.(2023)Qu, Gu, Xia, Li, Wang, and Huai}]{qu2023survey}
Xiaoye Qu, Yingjie Gu, Qingrong Xia, Zechang Li, Zhefeng Wang, and Baoxing Huai. 2023.
\newblock A survey on arabic named entity recognition: Past, recent advances, and future trends.
\newblock \emph{IEEE Transactions on Knowledge and Data Engineering}.

\bibitem[{Walker et~al.(2005)Walker, Strassel, Medero, and Maeda}]{walker2005ace}
Christopher Walker, Stephanie Strassel, Julie Medero, and Kazuaki Maeda. 2005.
\newblock Ace 2005 multilingual training corpus-linguistic data consortium.
\newblock \emph{URL: https://catalog. ldc. upenn. edu/LDC2006T06}.

\bibitem[{Wang et~al.(2022)Wang, Li, Meng, Zhang, Ouyang, Li, and Wang}]{knnner}
Shuhe Wang, Xiaoya Li, Yuxian Meng, Tianwei Zhang, Rongbin Ouyang, Jiwei Li, and Guoyin Wang. 2022.
\newblock $ k $ nn-ner: Named entity recognition with nearest neighbor search.
\newblock \emph{arXiv preprint arXiv:2203.17103}.

\bibitem[{Weischedel et~al.(2013)Weischedel, Palmer, Marcus, Hovy, Pradhan, Ramshaw, Xue, Taylor, Kaufman, Franchini et~al.}]{weischedel2013ontonotes}
Ralph Weischedel, Martha Palmer, Mitchell Marcus, Eduard Hovy, Sameer Pradhan, Lance Ramshaw, Nianwen Xue, Ann Taylor, Jeff Kaufman, Michelle Franchini, et~al. 2013.
\newblock Ontonotes release 5.0 ldc2013t19.
\newblock \emph{Linguistic Data Consortium, Philadelphia, PA}, 23:170.

\bibitem[{Zaratiana et~al.(2022)Zaratiana, Tomeh, Holat, and Charnois}]{span-based}
Urchade Zaratiana, Nadi Tomeh, Pierre Holat, and Thierry Charnois. 2022.
\newblock Named entity recognition as structured span prediction.
\newblock In \emph{Proceedings of the Workshop on Unimodal and Multimodal Induction of Linguistic Structures (UM-IoS)}, pages 1--10.

\end{thebibliography}

\appendix

\end{document}